%% file: main.tex
\documentclass[runningheads]{llncs}
\usepackage[T1]{fontenc}
\usepackage{graphicx,verbatim}
\input{preamble}

\begin{document}
\title{Information Maximization for Long-Tailed Semi-Supervised Domain Generalization}
\titlerunning{Information Maximization for Long-Tailed SSDG}
%

\author{Leo Fillioux\textsuperscript{1*}, Omprakash Chakraborty\textsuperscript{2}, Quentin Gopée\textsuperscript{1}, Pierre Marza\textsuperscript{1}, Paul-Henry Cournède\textsuperscript{1}, Stergios Christodoulidis\textsuperscript{1}, Maria Vakalopoulou\textsuperscript{1}, Ismail Ben Ayed\textsuperscript{2} and Jose Dolz\textsuperscript{2}}  
\authorrunning{L. Fillioux et al.}
\institute{\textsuperscript{1}Université Paris-Saclay, CentraleSupélec, Gustave Roussy, INSERM, CDSU, IHU PRISM, Gif-sur-Yvette\\
\textsuperscript{2}LIVIA, ILLS, ETS Montréal\\
    \email{leo.fillioux@centralesupelec.fr}}
  
\maketitle              

\input{sections/0_abstract}
\input{sections/1_introduction}
\input{sections/2_related_work}
\input{sections/3_methodology}
\input{sections/4_experiments}
\input{sections/5_conclusion}

\bibliographystyle{splncs04}
\bibliography{main}

\end{document}

%% file: preamble.tex
\usepackage[dvipsnames]{xcolor}
\usepackage{color, colortbl}
\usepackage{amsmath}
\usepackage{amsfonts}
\usepackage{graphicx,verbatim}
\usepackage{booktabs}
\usepackage{multirow}
\usepackage{subcaption}
\usepackage{caption}
\usepackage{dsfont}

\newcommand{\mypar}[1]{\noindent\textbf{#1}\quad}
\newcommand{\params}{\boldsymbol{\theta}}

\newcommand{\vy}{\mathbf{y}}
\newcommand{\vx}{\mathbf{x}}
\newcommand{\vp}{\mathbf{p}}
\newcommand{\vu}{\mathbf{u}}

\newcommand{\calX}{\mathcal{X}}
\newcommand{\calY}{\mathcal{Y}}
\newcommand{\calH}{\mathcal{H}}
\newcommand{\calD}{\mathcal{D}}

\definecolor{lightorange}{rgb}{0.98, 0.95, 0.92}
\newcommand{\our}{\cellcolor{lightorange}}

\definecolor{Better}{rgb}{0.18, 0.407, 0.266}
\definecolor{Worse}{rgb}{0.55, 0.22, 0.22}
\definecolor{Same}{rgb}{0.55, 0.55, 0.55}
\newcommand{\imp}[1]{$_{{\textbf{\textcolor{Better}{#1}}}}$}
\newcommand{\wor}[1]{$_{{\textbf{\textcolor{Worse}{#1}}}}$}

\definecolor{myblue}{RGB}{38, 145, 199}

\newcommand{\ours}{\textsc{IMaX}}

\newcommand\blfootnote[1]{%
  \begingroup
  \renewcommand\thefootnote{}\footnote{#1}%
  \addtocounter{footnote}{-1}%
  \endgroup
}

%% file: sections/0_abstract.tex
\begin{abstract}
    Semi-supervised domain generalization (SSDG) has recently emerged as an appealing alternative to tackle domain generalization when labeled data is scarce but unlabeled samples across domains are abundant. In this work, we identify an important limitation that hampers the deployment of state-of-the-art methods on more challenging but practical scenarios. In particular, state-of-the-art SSDG severely suffers in the presence of long-tailed class distributions, an arguably common situation in real-world settings. To alleviate this limitation, we propose \textbf{\ours{}}, a simple yet effective objective based on the well-known InfoMax principle adapted to the SSDG scenario, where the Mutual Information (MI) between the learned features and latent labels is maximized, constrained by the supervision from the labeled samples. Our formulation integrates an $\alpha$-entropic objective, which mitigates the class-balance bias encoded in the standard marginal entropy term of the MI, thereby better handling arbitrary class distributions. \ours{} can be seamlessly plugged into recent state-of-the-art SSDG, consistently enhancing their performance, as demonstrated empirically across two different image modalities.
    \blfootnote{\textsuperscript{*} Work done during an internship at ILLS.}
    \keywords{Domain generalization \and Semi-supervised learning.}
\end{abstract}

%% file: sections/1_introduction.tex
\section{Introduction}
\label{sec:introduction}

Deep learning approaches have dominated the computer vision field in the last decade, with unprecedented performances in a plethora of visual recognition problems \cite{krizhevsky2012imagenet}. However, these models are typically trained under the assumption that training and testing samples follow the same distribution. When distributional drifts violate this \textit{i.i.d.} (independent and identically distributed) assumption, their performance can degrade substantially \cite{hendrycks2021natural,recht2019imagenet}.

To address this problem, Domain Generalization (DG) aims to learn models that remain robust to unseen target domains \cite{li2020domain,ouyang2022causality,zhou2022domain}. In particular, in DG scenarios we assume that labeled data from multiple source domains is available, and the primary objective is to train a model that can adequately generalize to novel unseen domains. However, in real-world scenarios, such as healthcare, where labeling samples for all the domains may become impractical, the assumption that the data from multiple source domains are fully labeled is unrealistic, and generally unmet. Thus, to reduce the burden of the annotation process, semi-supervised domain generalization (SSDG) has been recently explored \cite{zhou2021stylematch,galappaththige2024towards}, which unifies the problems of semi-supervised learning and domain generalization. In this setting, each source domain has access to a few labeled samples and a larger amount of unlabeled data. Thus, beyond striving for cross-domain generalization, the model should achieve this objective efficiently with respect to labeled data requirements.

\begin{figure}[t]
\vspace{-1em}
    \centering
    \begin{subfigure}[t]{0.49\linewidth}
        \centering
        \raisebox{.13\height}{\includegraphics[width=\linewidth]{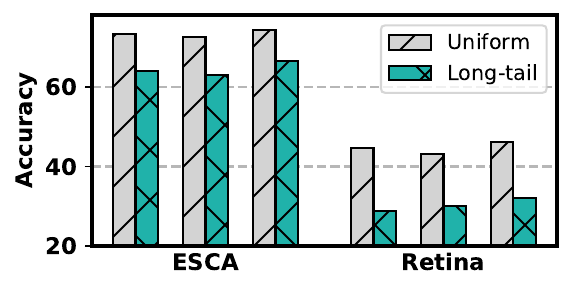}}
    \end{subfigure}
    \begin{subfigure}[t]{0.49\linewidth}
        \centering
        \includegraphics[width=\linewidth]{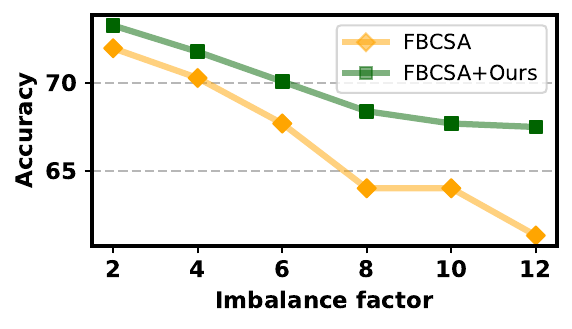}
    \end{subfigure}
    \caption{\textbf{Performance degradation under class imbalance.} Impact on the accuracy of having long-tailed class distribution in the training data using the FBCSA~\cite{galappaththige2024towards} trainer (\textit{left}). Impact of the imbalance factor $\gamma$ (see Section~\ref{subsec:experimental_setup}) on the performance of FBCSA and our proposed approach (\textit{right}).}
    \label{fig:motivation_plot}
    \vspace{-3mm}
\end{figure}

Very recently, \cite{galappaththige2024towards} exposed that several existing semi-supervised learning (SSL) approaches outperformed methods tailored to DG in the SSDG scenario. Based on these observations, authors proposed two plug-and-play strategies (i.e., FBCSA \cite{galappaththige2024towards} and DGWM \cite{galappaththige2025domain}) that can be integrated into different SSL methods, setting a novel state-of-the-art for this task. Nevertheless, after exploring the proposed frameworks we identified an important limitation that hampers their application in real-life scenarios. In particular, these approaches assume uniform class distributions across the different source domains. However, in real-world applications, many tasks involve inherently imbalanced data, \textit{e.g.}, rare diseases in medical imaging. Indeed, we empirically observed that if we assume \textit{long-tailed} class distributions, the performance of both FBCSA \cite{galappaththige2024towards} and DGWM \cite{galappaththige2025domain} is substantially degraded across several tasks (Fig.~\ref{fig:motivation_plot}), underscoring the importance of considering this more realistic scenario. 

In light of the observation presented above, in this paper we investigate the problem of domain generalization in the context of semi-supervised learning with class-imbalanced datasets, which combines the problems of domain generalization and label-efficient learning in the presence of class imbalance, a more realistic scenario than prior works \cite{galappaththige2024towards,galappaththige2025domain}. Thus, our contributions can be summarized as:
\renewcommand{\labelitemi}{\textbullet}
\begin{itemize}
    \item We introduce a more realistic SSDG setting where, in addition to the standard labeled and unlabeled data across multiple source domains, labeled samples are exposed to imbalanced class distributions. 
    \item We present \textbf{\ours{}}, an information theoretic approach to tackle this challenging yet realistic scenario. In particular, the proposed solution accommodates the specificities of this challenging setting by deriving a semi-supervised view of the standard Mutual Information (MI). 
    \item To further account for class imbalance, we replace the less flexible marginal entropy term in the modified MI by a learning objective derived from Tsallis divergences~\cite{tsallis1988}, which tolerates better variations on class distributions.
    \item Experiments across several datasets, SSL and SSDG approaches demonstrate that \ours{} consistently enhances existing methods in the studied scenario, showcasing its model-agnosticity and \textit{plug‑and‑play} versatility. 
\end{itemize}

%% file: sections/2_related_work.tex
\section{Related Work}
\label{sec:related_work}

\noindent \textbf{Semi-supervised Domain Generalization (SSDG)} focuses on improving generalization across domains when labeled data is scarce but substantial unlabeled data is accessible. Within the broader computer vision community, state-of-the-art SSDG \cite{galappaththige2024towards,galappaththige2025domain} leverage semi-supervised learning. These methods consist in generating pseudo-labels from weakly augmented samples, and training the model to enforce that predictions from strongly augmented versions to align with these pseudo-labels. In particular, FixMatch \cite{fixmatch}, FreeMatch \cite{wang2023freematch}, and StyleMatch \cite{zhou2021stylematch} are typically employed as SSL frameworks in SSDG \cite{galappaththige2024towards,galappaththige2025domain}. The SSDG setting is particularly well suited to medical imaging, where acquisition conditions can vary widely. In this scenario, different approaches have been developed specifically for medical imaging problems \cite{zhang2022semi,tissir2025style,song2025dual}. However, these techniques typically rely on restrictive, modality‑dependent assumptions, making them applicable only to a subset of medical imaging modalities. For example, \cite{zhang2022semi} assumes that the distribution shift results in a modification of the hue saturation and contrast, while \cite{song2025dual} assumes that structural similarity between classes is preserved across domains. Both conditions do not hold consistently across diverse and real clinical settings, e.g., anatomical variability and pathology‑dependent morphology can substantially alter structural patterns.

\mypar{Mutual information (MI) in deep learning.} Our approach is grounded in the well‑established \textit{InfoMax} principle \cite{Linsker1988_SelfOrganizationInAPerceptualNetwork}, which advocates maximizing the MI between a system’s inputs and outputs. Several variants of this general principle, which accommodate the specific requirements of different scenarios, have been recently used in machine learning and computer vision tasks, including deep clustering \cite{jabi2019deep,krause2010discriminative,ohl2022generalised}, few-shot learning \cite{boudiaf2021few,boudiaf2020information,hajimiri2023strong}, representation learning \cite{bachman2019learning,hjelm2019learning,kemertas2020rankmi}, or generalized category discovery \cite{chiaroni2023parametric}. To the best of our knowledge, however, addressing SSDG from an information-theoretic standpoint remains unexplored.

%% file: sections/3_methodology.tex
\section{Methodology}
\label{sec:methodology}

\noindent \textbf{Problem definition and notation.} We now formally define the semi-supervised domain generalization (SSDG) problem. Let $\calX$ and $\calY$ define the input and label space, respectively. A \textit{domain} is defined by a joint probability distribution $P_{XY}$ over $\calX\times \calY$. In this scenario, we only consider distribution shifts in $P_X$, while assuming $P_Y$ remains the same, i.e., all domains share the same label space. We have access to $D$ distinct but related domain sources $\mathcal{S}=\{S_d\}_{d=1}^D$, each associated with a joint distribution $P^{(d)}_{XY}$. In SSDG, only a small portion of the source data have labels, while many instances are unlabeled, for which we can only access the empirical marginal distribution $P_X$. For each source domain $S_d$, the labeled subset is defined as $S^{(d)}_L=\{(\vx^{(d)}_i, \vy^{(d)}_i)\}^{m_L^{(d)}}_{i=1}$, where $(\vx^{(d)}, \vy^{(d)})\sim P^{(d)}_{XY}$, and the unlabeled subset as $S^{(d)}_U=\{\vx^{(d)}_i\}_{i=1}^{m_U^{(d)}}$, with $\vx^{(d)}\sim P^{(d)}_X$, which denotes the marginal distribution of $P^{(d)}_{XY}$ over $\calX$. The size of the unlabeled dataset is typically much larger than that of the labeled data, i.e., $m_U^{(d)} = \vert S_U^{(d)}\vert \gg \vert S_L^{(d)}\vert = m_L^{(d)}$.

The goal in SSDG is to learn a domain-generalizable model, parameterized by $\params$, using both labeled and unlabeled source data, which outputs a softmax prediction vector $\vp_i = (p_{ik})_{1\leq k \leq K}$, with $K$ denoting the number of classes. At test time, the model is directly deployed in an unseen target domain $T=\{\vx^*\}$ with $\vx^*\sim P^*_X$. The target domain differs from any source domain used for training, i.e., $ P^*_{X} \neq P^{(d)}_{X}$ for all $d \in \{ 1, \dots, D \}$.

\noindent \textbf{Mutual information.} The mutual information (MI) between the labels and the input images can be defined as follows:
\begin{equation}
\label{eq:mutual_info}
    I(Y;X) = \mathcal{H}(Y) - \mathcal{H}(Y|X),
\end{equation}
where $\mathcal{H}(Y)$ denotes the entropy of the marginal distributions $\pi_k = \mathbb{P}(Y=k;\params)$, and $\mathcal{H}(Y|X)$ refers to the entropy of the conditional probability distribution $\mathbb{P}(Y|X; \params)$, \textit{i.e.}, the remaining uncertainty in $Y$ when $X$ is known. The marginal distribution can be approximated by the soft proportion of samples within each class $k \in \{1, \ldots K\}$ across both the set of all labeled ($\calD_L$) and unlabeled ($\calD_U$) samples $\calD=\{\calD_L, \calD_U\}$, as $\pi_k \approx \frac{1}{|\calD|} \sum_{i \in \calD} p_{ik}$.

\noindent \textbf{Our method.} While it is common in the literature to maximize the unsupervised mutual information in Eq. (\ref{eq:mutual_info}) (e.g., often defined over unlabeled samples \cite{boudiaf2020information,krause2010discriminative}), we accommodate this term to our current scenario, where a few labeled samples are available. More concretely, we integrate explicit supervision constraints on the conditional probabilities $\vp_i$ of the samples within the labeled set. The proposed constrained information maximization can be formally defined as:
\begin{equation}
        \max_{\params} \, \mathcal{H}(Y) - \mathcal{H}(Y | X) \quad \mbox{s.t.} \quad \vy_i=\vp_i \quad \forall \vx_i \in \calD_L
    \label{eq_constrained_mutual_info}
\end{equation}
Integrating the equality constraints above into the MI, the objective becomes: 
\begin{equation}
    \begin{aligned}
        \min_{\params} \, \sum_{k=1}^K \pi_k \log \pi_k 
    - \frac{1}{|\calD|} \sum_{i \in \calD} \sum_{k=1}^K q_{ik} \log p_{ik},
    \end{aligned}
    \label{eq_auxiliary}
\end{equation}
with $q_{ik}=y_{ik}$ if $\vx_i \in \calD_L$ and $q_{ik} = p_{ik}$ otherwise. Further decomposing this the equation above, we can obtain the following objective:
\begin{equation}
    \begin{aligned}
\min_{\params} \quad \sum_{k=1}^K \pi_k \log \pi_k -\frac{1}{|\calD_L|} \sum_{i \in \calD_L} \sum_{k=1}^K y_{ik} \log p_{ik}
      - \frac{1}{|\calD_U|} \sum_{i \in \calD_U} \sum_{k=1}^K p_{ik} \log p_{ik}
    \end{aligned}
    \label{eq:MI_three_terms}
\end{equation}

\noindent \textit{Semi-Supervised Mutual Information.} The last term in Eq.~(\ref{eq:MI_three_terms}) represents the conditional Shannon entropy, and is an unsupervised objective that  encourages implicitly confident predictions, as it pushes them toward one vertex of the simplex. Instead of simply optimizing the third objective, i.e., minimizing the entropy of the predictions, we leverage the concepts of \textit{consistency regularization} and \textit{pseudo-labeling}, widely used in the semi-supervised learning literature \cite{fixmatch,wang2023freematch,zhou2021stylematch}. More concretely, all unlabeled images undergo an augmentation step, where weak and strong transformations of each image are performed. Then, pseudo-labels derived from the predictions on weak image transformations are used to guide the predictions of their strongly augmented counterparts \cite{fixmatch}. Formally, given a \textit{weakly}-augmented version of unlabeled image $i$, we obtain its softmax prediction vector $\vp_i$, from which we can compute its pseudo-label $\hat{y}_i=\arg \max (\vp_i)$\footnote{Pseudo-label $\hat{y}_i$ is transformed into a one-hot encoded vector $\hat{\vy}_i \in \{0,1\}^K$ to be used in (\ref{eq:pseudo_ce}) and (\ref{eq:MI_SSL}).}. Thus, for image $i$, we can define the pseudo cross-entropy as:
\begin{equation}
\label{eq:pseudo_ce}
    - \sum_{k=1}^K \mathds{1}(\max(\vp_i) \geq \tau) \hat{y}_{ik} \log p_{ik}^{\mathcal{A}},
\end{equation}
\noindent where $\vp_i^{\mathcal{A}}$ denotes the softmax prediction vector of the strongly augmented image, i.e., $\vp_i^{\mathcal{A}}=f_{\params}(\mathcal{A}(\vx_i))$, and $\tau$ a scalar hyperparameter denoting the threshold above which a pseudo-label should be considered. Plugging this term into Eq.~(\ref{eq:MI_three_terms}), we have the following SSL view of the MI: 
\begin{equation}
    \begin{aligned}
\min_{\params} \quad \underbrace{\sum_{k=1}^K \pi_k \log \pi_k}_{ -\calH(Y): \text{ marginal entropy}} & \underbrace{-\frac{1}{|\calD_L|} \sum_{i \in \calD_L} \sum_{k=1}^K y_{ik} \log p_{ik}}_{\calH (Y|X_L) \text{: cross-entropy}} \\
      &- \underbrace{\frac{1}{|\calD_U|} \sum_{i \in \calD_U} \sum_{k=1}^K \mathbb{I}(\max(\vp_i) \geq \tau) \hat{y}_{ik} \log p_{ik}^{\mathcal{A}}}_{\calH (\hat{Y}|X_U)\text{: pseudo cross-entropy}}
    \end{aligned}
    \label{eq:MI_SSL}
\end{equation}

\noindent \textit{Adapting to imbalanced scenarios.} While the marginal entropy term $\mathcal{H}(Y)$ in Eq. (\ref{eq:MI_SSL}) is of paramount importance to prevent trivial arbitrary solutions \cite{boudiaf2020information}, it pushes the marginal distribution towards the uniform,  
thereby encoding a strong bias towards balanced partitions. Nevertheless, as shown in our empirical observations (Fig. \ref{fig:motivation_plot}), and stated in our motivations, assuming uniform class partitions is unrealistic in real-world problems. Inspired by \cite{veilleux2021realistic}, we propose a generalization of the semi-supervised mutual-information loss in Eq. (\ref{eq:MI_SSL}), based on $\alpha$-divergences, which can better accommodate variations in class distributions. In particular, we focus on Tsallis’~\cite{tsallis1988} formulation of the $\alpha$-divergences. Indeed, if we write the marginal entropy as an explicit Kullback-Leibler (KL) divergence ($\calH(Y)=\log (K) - \calD_{KL}(\vp||\vu_K)$, where $\vu_K$ is the probability distribution for the uniform distribution with $K$ classes), it generalizes to Tsallis $\alpha$-entropy:
\begin{align}
\label{eq:alpha-ent}
    \calH_{\alpha}(\vp) &= \log_{\alpha}(K) - K^{1-\alpha} \calD_{\alpha}(\vp \| \vu_K) = \frac{1}{\alpha - 1} \left( 1 - \sum_{k} p^{\alpha}_k \right)
\end{align}
A derivation of the above equation is provided in Appendix in \cite{veilleux2021realistic}.

\noindent \textit{Final objective.} Our final objective is then formally defined as:
\begin{equation}
   \min_{\params} \quad -\calH_{\alpha}(Y)+\calH(Y|X_L)+\calH(\hat{Y}|X_U),
   \label{eq:our_objective}
\end{equation}
\noindent where $\calH_{\alpha}(Y)$ is the $\alpha\textsc{-}$entropy defined in Eq. (\ref{eq:alpha-ent}). The role of each term in the objective presented above is:

\begin{itemize}
    \item Standard cross-entropy on labeled samples $\calH(Y|X_L)$ could be seen as the \textit{penalty} function employed to enforce the constraints $\vy_i=\vp_i, \, \forall \vx_i \in \calD_L$, thereby avoiding the need to explicitly enforce the equality constraints in Eq.~(\ref{eq_constrained_mutual_info}). Thus, this term basically encourages that, for labeled samples, network predictions align with their corresponding class labels.
    \item Pseudo cross-entropy for unlabeled samples $\calH(\hat{Y}|X_U)$ acts as a pseudo supervisory signal on unlabeled samples, following standard SSL practices \cite{fixmatch,wang2023freematch,zhou2021stylematch}. It replaces the conditional entropy $\calH(Y|X)$, as its optima may easily lead to degenerate solutions, mapping all samples to a single arbitrary class. Using pseudo-labels, as well as a mechanism to filter only confident ones, prevents these degenerate solutions, while also avoiding the special care required during optimization of $\calH(Y|X)$ \cite{boudiaf2020information,dhillonbaseline}.
    \item $\calH_{\alpha}(Y)$ acts as a label marginal regularization, offering a more relaxed version of the standard label marginal entropy in MI, which encourages 
    a near-uniform class distribution. By relaxing the strictly uniform constraint, 
    $\calH_{\alpha}(Y)$ tolerates class distributions further from the uniform, which makes the proposed solution more suitable for imbalanced scenarios.
\end{itemize}

%% file: sections/4_experiments.tex
\section{Experiments}
\label{sec:experiments}

\subsection{Experimental setup}
\label{subsec:experimental_setup}
\noindent \textbf{Datasets.} We explore 
two different tasks. 
\textbf{Histology:} ESCA~\cite{tolkach2023artificial}, a histo-pathology patch-level classification dataset with images from 11 classes across four different hospitals, each treated as a separate domain.
\textbf{Ophthalmology:} we focus on diabetic retinopathy (DR) grading, with five different grades (no DR, mild DR, moderate DR, severe DR, and proliferative DR). We use four different datasets, each simulating a separate domain: 
Messidor-2~\cite{Etienne2014,Krause2018}, IDRiD~\cite{Porwal2020}, Paraguay~\cite{PARAGUAY}, and APTOS~\cite{aptos}. Labels are standardized following~\cite{FLAIR}.

\noindent \textbf{Baselines.} Since our objective is to design an approach that does not rely on impractical assumptions, and remains broadly applicable, we focus our comparisons on \textit{assumption‑free} state-of-the-art SSDG methods, such as FBCSA~\cite{galappaththige2024towards} and DGWM \cite{galappaththige2025domain}. Furthermore, following the recent SSDG literature \cite{galappaththige2024towards,galappaththige2025domain}, we leverage three common SSL methods: FixMatch~\cite{fixmatch}, FreeMatch~\cite{wang2023freematch}, and StyleMatch~\cite{zhou2021stylematch}.

\noindent \textbf{Implementation details.} In each task, each experiment consists of 20 runs, corresponding to four setups (train on three domains and test on the fourth) and five seeds for each setup, reporting the average metric across runs. We train for 20 epochs using SGD. Based on the validation set, we empirically set $\alpha=1.5$ (ESCA dataset), and $\alpha=2$ (Retina dataset).
\textit{Enforcing imbalance:} Given $K$ classes, considering $m_L$ labeled samples per class in each domain, we take $m_L\cdot K$ samples per domain, and enforce an exponentially decaying number of samples per class, shuffling the order of the classes differently for each seed. The hyperparameter $\gamma$ (set to $\gamma=10$) tunes the decay, and therefore the imbalance. 

\subsection{Results}
\noindent \textbf{Main results.} In this section we evaluate the performance of \ours{} when coupled with existing SSDG methods. In addition to the existing FBCSA~\cite{galappaththige2024towards} and DGWM \cite{galappaththige2025domain}, we refer to \textit{``Baseline''} as the direct use of SSL methods, without any SSDG-specific component. Table~\ref{tab:main} reports the average accuracy for the ESCA and Retina datasets considering $m_L\in\{5,10\}$ per class and per domain, across three different frameworks and three SSL methods. We observe that \textbf{\ours{} consistently improves performance across all but one setting}. Notably, improvements are most pronounced in the low-label regime (up to \textbf{+7.3\%} when $m_L=5$ \textit{vs.} \textbf{+5.8\%} with $m_L=10$), highlighting the effectiveness of our approach in scenarios where annotated data is scarce. These results underscore the practical relevance of \ours{}, as it enhances generalization of existing SSDG methods under real-world constraints, such as data-scarcity and long-tailed distributions.

\input{tables/main_results}

\noindent \textbf{Sensitivity to varying distributions (Fig.~\ref{fig:motivation_plot}).} We now assess robustness under increasingly extreme long‑tailed distributions, revealing how quickly existing SSDG methods deteriorate as imbalance grows. Fig. \ref{fig:motivation_plot} (\textit{right}) depicts the results on the ESCA dataset with the FBCSA strategy, with and without applying the proposed \ours{}. The results show that, while both models see their performance decrease as data imbalance increases, \textbf{degradation without \ours{} is significantly more severe}, particularly for more extreme imbalance.

\noindent \textbf{Impact of the proposed learning objectives (Tab.~\ref{tab:uniform}).} Integrating the proposed semi-supervised MI objective (Eq. \ref{eq:MI_SSL}) into the baseline already yields substantial gains (\textbf{+5.0\%} and \textbf{+3.4\%} for $m_L \in \{5,10\}$). We note that, although this formulation improves over the baseline, it remains a rigid case of our $\alpha$-based objective, since Eq.~\ref{eq:MI_SSL} is a particular case of the more general objective in Eq.~\ref{eq:our_objective}, when $\alpha=1$. Relaxing this term with the more flexible $\calH_{\alpha}$ term yields further gains (\textbf{+5.0\%} and \textbf{+3.4\%}), demonstrating the effectiveness of \ours{}.

\input{tables/ablation}

\begin{figure}
\vspace{-0.5em}
    \centering
    \begin{subfigure}[t]{0.4\linewidth}
        \centering
        \includegraphics[width=\linewidth]{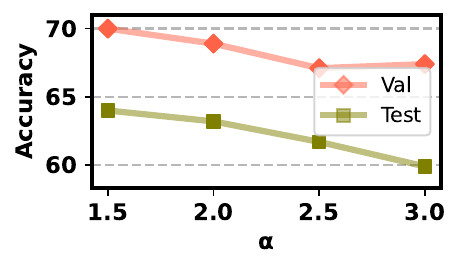}
    \end{subfigure}
    \begin{subfigure}[t]{0.4\linewidth}
        \centering
        \includegraphics[width=\linewidth]{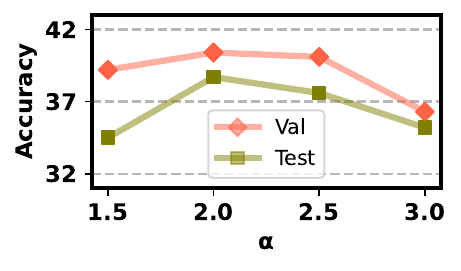}
    \end{subfigure}
    \vspace{-1.5em}
    \caption{\textbf{Ablation on the $\boldsymbol\alpha$ parameter.} Impact on the $\alpha$ parameter of $\calH_{\alpha}$ on the accuracy for the ESCA (\textit{left}) Retina (\textit{right}) datasets under long-tailed distribution. Results are for FreeMatch on the SSL Baseline setting.}
    \label{fig:ablation_alpha}
    \vspace{-2em}
\end{figure}

\noindent \textbf{Sensitivity to $\boldsymbol{\alpha}$.} Fig.~\ref{fig:ablation_alpha} empirically validates our choices for $\alpha$. Indeed, while performance varies with $\alpha$, the validation and test accuracies exhibit highly consistent trends. This alignment indicates that the value selected on the validation set transfers reliably to the test domain, ensuring that $\alpha$ can be tuned in a stable and predictable manner.

%% file: tables/main_results.tex
\begin{table}[h!]
\vspace{-2em}
    \centering
    \scriptsize
    \caption{\textbf{Performance in the long-tailed SSDG scenario.}
    We report [standard\:/\:\textbf{+\ours{}}], where \textbf{+\ours{}} denotes the results obtained by our approach when applied on each SSDG method and across the three SSL strategies.}
    \label{tab:main}
    \vspace{0.4em}
    \begin{tabular}{
    >{\centering\arraybackslash}m{0.4cm}   
    >{\raggedright\arraybackslash}m{2.9cm}   
    >{\centering\arraybackslash}m{2.0cm}     
    >{\centering\arraybackslash}m{2.0cm}     
    >{\centering\arraybackslash}m{2.0cm}     
    >{\centering\arraybackslash}m{2.0cm}     
}
    \toprule
        & & \multicolumn{2}{c}{\textbf{ESCA}} & \multicolumn{2}{c}{\textbf{Retina}} \\
        \cmidrule(lr){3-4} \cmidrule(lr){5-6}
        & \textbf{Method} & $m_L=5$ & $m_L=10$ & $m_L=5$ & $m_L=10$ \\
        \midrule
        \multirow{4}{*}{\rotatebox[]{90}{Baseline}} & 
        FixMatch/\bf +\ours{} &
            $61.0\:/\:\bf 68.3$\imp{+7.3} &
            $69.2\:/\:\bf73.1$\imp{+3.9} &
            $31.4\:/\:\bf38.4$\imp{+7.0} &
            $34.4\:/\:\bf40.2$\imp{+5.8} \\
        \cmidrule{2-6}
        & FreeMatch/\bf +\ours{} & 
            $59.3\:/\:\bf63.2$\imp{+3.9} &
            $67.5\:/\:\bf71.0$\imp{+3.5} &
            $36.0\:/\:\bf38.7$\imp{+2.7} &
            $34.0\:/\:\bf37.0$\imp{+3.0} \\
        \cmidrule{2-6}
        & StyleMatch/\bf +\ours{} &
            $60.4\:/\:\bf67.7$\imp{+7.3} &
            $69.0\:/\:\bf73.4$\imp{+4.4} &
            $30.9\:/\:\bf36.0$\imp{+5.1} &
            $34.6\:/\:\bf38.4$\imp{+3.8} \\
    \midrule
        \multirow{4}{*}{\rotatebox[]{90}{FBCSA}} & 
        FixMatch/\bf +\ours{}  &
            $64.0\:/\:\bf67.7$\imp{+3.7} & 
            $72.2\:/\:\bf73.2$\imp{+1.0} &
            $28.8\:/\:\bf32.3$\imp{+3.5} &
            $\textbf{32.0}\:/\:29.5$\wor{-2.5} \\
        \cmidrule{2-6}
        & FreeMatch/\bf +\ours{}  &
            $63.1\:/\:\bf65.5$\imp{+2.4} &
            $71.6\:/\:\bf72.4$\imp{+0.8} &
            $30.1\:/\:\bf31.1$\imp{+1.0} &
            $27.9\:/\:\bf29.8$\imp{+1.9} \\
        \cmidrule{2-6}
        & StyleMatch/\bf +\ours{} &
            $66.5\:/\:\bf69.9$\imp{+3.4} &
            $73.3\:/\:\bf75.3$\imp{+2.0} &
            $32.1\:/\:\bf35.6$\imp{+3.5} &
            $30.0\:/\:\bf32.0$\imp{+2.0} \\
    \midrule
        \multirow{4}{*}{\rotatebox[]{90}{DGWM}} &
        FixMatch/\bf +\ours{} &
            $63.2\:/\:\bf68.8$\imp{+5.6} &
            $71.5\:/\:\bf74.3$\imp{+2.8} &
            $26.1\:/\:\bf31.8$\imp{+5.7} &
            $31.4\:/\:\bf33.2$\imp{+1.8} \\
        \cmidrule{2-6}
        & FreeMatch/\bf +\ours{} &
            $62.2\:/\:\bf66.1$\imp{+3.9} &
            $71.1\:/\:\bf72.6$\imp{+1.5} &
            $30.6\:/\:\bf30.8$\imp{+0.2} &
            $32.3\:/\:\bf33.5$\imp{+1.2} \\
        \cmidrule{2-6}
        & StyleMatch/\bf +\ours{} &
            $63.9\:/\:\bf69.3$\imp{+5.4} &
            $71.7\:/\:\bf73.8$\imp{+2.1} &
            $27.5\:/\:\bf31.3$\imp{+3.8} &
            $31.8\:/\:\bf33.7$\imp{+1.9} \\
    \bottomrule
    \end{tabular}
    \vspace{-1em}
\end{table}

%% file: tables/ablation.tex
\begin{table}[h!]
\vspace{-2em}
\scriptsize
    \centering
    \caption{\textbf{Ablation on the proposed learning objectives,} where SSL baseline can be defined as $\mathcal{H}(Y|X_L)+\mathcal{H}(\hat{Y}|X_U)$. Comparison on the ESCA dataset.}
    \label{tab:uniform}
    \vspace{0.4em}
    \begin{tabular}{
    >{\raggedright\arraybackslash}m{5.9cm}   
    >{\centering\arraybackslash}m{1.9cm}     
    >{\centering\arraybackslash}m{1.9cm}     
}
    \toprule
        \textbf{Loss} & \textbf{$m_L=5$} & \textbf{$m_L=10$} \\
        \midrule
        SSL baseline & 61.0 & 69.2 \\
        \our SSL baseline + $(-\mathcal{H}(Y))$ (Eq. \ref{eq:MI_SSL}) & \our 66.0\imp{+5.0} & \our 72.6\imp{+3.4} \\
        \our SSL baseline + $(-\calH_{\alpha}(Y))$ (Eq. \ref{eq:our_objective}) = \textbf{\ours{}}
         & \our \bf 68.3\imp{+7.3} & \our \bf 73.1\imp{+3.9} \\
    \bottomrule
    \end{tabular}
    \vspace{-1em}
\end{table}

%% file: sections/5_conclusion.tex
\vspace{-1em}
\section{Conclusion}
\label{sec:conclusion}
We introduce the long-tailed SSDG setting, a realistic scenario which has proven to be particularly challenging for current SSDG methods. To accommodate to the challenges of long-tailed SSDG, we propose \ours{}, which leverages the InfoMax principle, and introduces a specific learning objective that acounts for class imbalance. Furthermore, \ours{} is model-agnostic and plug-and-play, enabling seamless integration with SoTA SSDG frameworks based on SSL, a popular strategy in the literature, showing consistent improvement across settings.

\mypar{Acknowledgements.} This work has benefited from state financial aid, managed by the Agence Nationale de Recherche under the investment program integrated into France 2030, project reference ANR-21-RHUS-0003. We acknowledge the support of the Natural Sciences and Engineering Research Council of Canada (NSERC). This work was granted access to the HPC resources of IDRIS under the allocation 2025-AD011014802R1 made by GENCI. The author gratefully acknowledges support from ILLS during the internship in which this work was conducted.